\documentclass[journal]{elsarticle}
\pdfoutput=1 
\usepackage{lineno,hyperref}
\modulolinenumbers[5]
\usepackage[numbers]{natbib}
\usepackage{amsmath}
\usepackage{multirow}
\usepackage{url}
\usepackage{array}
\usepackage{eufrak}
\usepackage{subfigure}
\usepackage{graphicx}
\usepackage{ulem}
\usepackage{marvosym}

\begin{document}

\begin{frontmatter}

\title{Decomposing Word Embedding with the Capsule Network}

\author[mymainaddress]{Xin Liu\fnref{myfootnote}}
\fntext[myfootnote]{orcid=https://orcid.org/0000-0003-2802-594X}
\ead{hit.liuxin@gmail.com}

\author[mymainaddress]{Qingcai Chen\corref{mycorrespondingauthor}}
\cortext[mycorrespondingauthor]{Corresponding author}
\ead{qingcai.chen@gmail.com}

\author[mysecondaryaddress]{Yan Liu}
\ead{csyliu@comp.polyu.edu.hk}

\author[mymainaddress]{Joanna Siebert}
\ead{joannasiebert@yahoo.com}

\author[mymainaddress]{Baotian Hu}
\ead{baotiannlp@gmail.com}

\author[mymainaddress]{Xiangping Wu}
\ead{wxpleduole@gmail.com}

\author[mymainaddress]{Buzhou Tang}
\ead{tangbuzhou@gmail.com}


\address[mymainaddress]{Harbin Institute of Technology, Shenzhen}
\address[mysecondaryaddress]{Department of Computing, The Hong Kong Polytechnic University, Hong Kong}

\begin{abstract}
Word sense disambiguation tries to learn the appropriate sense of an ambiguous word in a given context. 
The existing pre-trained language methods and the methods based on multi-embeddings of word did not explore the power of the unsupervised word embedding sufficiently. 

In this paper, we discuss a capsule network-based approach, taking advantage of capsule’s potential for recognizing highly overlapping features and dealing with segmentation. We propose a {\bf{Cap}}sule network-based method to {\bf{Dec}}ompose the unsupervised word {\bf{E}}mbedding of an ambiguous word in{\bf{to}} context specific {\bf{S}}ense embedding, called CapsDecE2S. 
In this approach, the unsupervised ambiguous embedding is fed into capsule network to produce its multiple morpheme-like vectors, which are defined as the basic semantic language units of meaning. 
With attention operations, CapsDecE2S integrates the word context to reconstruct the multiple morpheme-like vectors into the context-specific sense embedding. To train CapsDecE2S, we propose a sense matching training method. In this method, we convert the sense learning into a binary classification that explicitly learns the relation between senses by the label of matching and non-matching. The CapsDecE2S was experimentally evaluated on two sense learning tasks, i.e., word in context and word sense disambiguation. Results on two public corpora Word-in-Context and English all-words Word Sense Disambiguation show that, the CapsDecE2S model achieves the new state-of-the-art for the word in context and word sense disambiguation tasks.
\end{abstract}

\begin{keyword}
Word sense learning\sep Capsule network \sep Word-in-Context \sep English all-words Word Sense Disambiguation

\end{keyword}

\end{frontmatter}


\section{Introduction}
Word meanings are determined by their contexts \cite{Feng2018Geometric}. It is a generally followed principle by unsupervised word embedding approaches \citep{collobert2008unified}, e.g., Word2vec~\cite{mikolov2013distributed} and GloVe~\citep{pennington2014glove}.
In these approaches, the words with similar semantic roles are mapped into nearby points in the embedding space. They successfully capture the word semantic properties.
However, there is still an important issue left, i.e., distinguishing between the multiple senses of an ambiguous word whose intended meaning is in some way unclear to the reader because of its multiple interpretations or definitions.

The ambiguous words usually present multiple senses in various contexts.
Many approaches for learning the appropriate sense of an ambiguous word in a given context have been proposed.
The pre-trained language models, e.g., ELMo~\cite{peters2018deep}, OPENAI GPT~\cite{GPT}, and BERT~\cite{devlin2019bert}, are popular methods for learning the word sense representation dynamically.
The word embeddings in ELMo are learned functions of the internal states of a deep bidirectional language model~\cite{peters2018deep}. The OPENAI GPT uses a left-to-right Transformer and is trained on the BooksCorpus (800M words)~\cite{GPT} while BERT uses the pre-trained language tasks to learn word representations with the transformer language model~\cite{devlin2019bert}. 
These methods output contextual embeddings that infer different representations induced by arbitrarily long contexts.
They have had a major impact on driving progress on downstream tasks ~\cite{Daniel2019Language}. However, no explicit sense label is involved in these methods.

The multi-embeddings of an ambiguous word are another popular solution for sense learning.
The ambiguous word is usually represented by multiple embeddings and each embedding corresponds to one of its senses.
Sense graphs~\cite{jauhar2015ontologically, pelevina2016making}, bilingual resources~\cite{reisinger2010multi,guo2014learning,neelakantan2014efficient,ettinger2016retrofitting}, and semantic network ~\cite{agirre2014random,moro2015semeval,mancini2017embedding,pasini2018two} are widely used for learning multiple embeddings.
\cite{jauhar2015ontologically} proposed to apply graph smoothing and predictive maximum likelihood models to learn
senses grounded in a specified ontology.
\cite{ettinger2016retrofitting} proposed to retrofit sense-specific word vectors using parallel text.
\cite{pilehvar2016conflated} extracted semantically related words from WordNet and computed the sense embeddings in turn.

The multi-embedding based methods usually require well-organized knowledge base, whose scale is usually smaller than that for unsupervised word embedding learning.
Then, a natural question emerges: can we learn some information for the proper word sense based on the unsupervised word embedding?
For example, in "$\mathcal{S}_1$:Which fruit contains more vitamin C, {\bf \textit{apple}} or strawberry " and "$\mathcal{S}_2$: {\bf\textit{Apple}} is about to release iPhone X",
the embedding "apple" gives higher similarities to the words related to one of its senses than others ("strawberry" and "iPhone", respectively) 
This phenomenon indicates that we may infer some exact sense information from the unsupervised word embedding~\citep{huang2012improving}.

However, the aforementioned example also shows that the unsupervised word embedding of an ambiguous word contains some redundant information from other senses.
Then an urgent problem is how to extract the useful information from the unsupervised embedding that simultaneously contains other senses.
For this problem, we believe that using the capsule network could be a solution.
The capsule network was first proposed by~\cite{sabour2017dynamic} for digital image recognition task, 
and it shows potential for recognizing highly overlapping features~\cite{sabour2017dynamic,sabour2018matrix}.
In capsule network, a capsule is a group of neurons that represent the instantiation parameters of the entity~\cite{sabour2017dynamic}, which means that a capsule is a vector. 
Capsules are also very good for dealing with segmentation.
A lower-level capsule prefers to hand out its output to higher-level capsules and the higher-level capsules select information from the lower-level capsules to represent more complex entities~\cite{sabour2017dynamic}. 
At each location in the feature, there is at most one instance of the type of entity that a capsule represents~\cite{sabour2017dynamic}.
In other words, each capsule represents one kind of entity which contains the unique information. 
The ambiguous word embedding is just like the highly overlapping feature and can take full advantage of the capsule network approach.
When the embedding goes through the capsule network, the capsule at one
level attends to some active capsules at the level below and ignores others.
Finally, the higher-level capsules contains the unique information, and the capsules are just like the morphemes since each represents one basic semantic unit of meaning.
The procedure acts like decomposing an ambiguous word embedding into multiple morpheme-like vectors.
Even though the capsule network has been widely used for image recognition, we are unaware of any capsule network based approaches for word sense disambiguation. We believe that capsule network can benefit the word sense disambiguation, especially decomposing the unsupervised word embedding.

However, applying the capsule network to embedding decomposing is not a trivial task. It is challenging how to allocate the weights for combining the decomposed morpheme-like vectors.
Following the general principle that word meaning is determined by its context~\cite{Feng2018Geometric}, a context attention operation can be used compute the weights.
Attention is the widely used mechanism to draw global dependencies between input and output~\cite{vaswani2017attention}.
Taking the context as input and each decomposed morpheme-like vector as output, the attention weights can be easily obtained.
By combining these morpheme-like vectors with different weights, we can reach different senses of the same word.

For word sense learning, some approaches learn the exact sense by a classification training with the senses as the labels~\cite{luo-etal-2018-leveraging,luo2018incorporating,raganato2017neural}. Some other works proposed to learn the word sense by k-nearest neighbor cluster training~\cite{Daniel2019Language,pasini-etal-2020-clubert,scarlini2020sensembert}. However, there are large amount of labels in the first group of methods and it is hard to train a model with so many labels. On the other hand, the cluster training-based methods may face the problem of inexact senses when clustering. Besides, both methods cannot solve the sense zero-shot problem in training and test set.  
Inspired by the recent works~\cite{HuangGlossBERT,kumar-etal-2019-zero,pilehvar2018wic}, we propose a flexible training method, called word sense matching training, which converts the problem to a binary classification by judging the matching state of two senses.
With this training method, though the same sense of the ambiguous word may cross different sentences, we still learn the appropriate sense representation. Following~\cite{HuangGlossBERT}, 
the gloss of word sense can be converted into a sentence with the form of "word:gloss", which provides the gold standard for matching. Besides, since the gloss covers all the senses in WordNet, we can match almost all senses and obstacle the problem of zero-shot. 
On one hand, unlike the previous classification training, the word sense matching method only outputs the matching label of two senses. On the other hand, the method uses the explicit sense as the gold standard for training and the zero-shot could be solved by matching the glosses from WordNet~\cite{kumar-etal-2019-zero}.

Our main contributions are summarized as follows.
1) We propose an embedding decomposing method with the capsule network. The capsule network decomposes the unsupervised word embedding into multiple morpheme-like vectors.
2) We propose a novel framework for merging morpheme-like vectors by contextual attention, which enables the dynamic generation of context specific sense representation. The context specific sense representation maintains the appropriate sense of an ambiguous word in a given context.
3) We propose a new training method called word sense matching for word sense learning, which integrates the gold sense label into the training procedure, and achieves the new state-of-the-art on word in context task and word sense disambiguation task, respectively.

\section{Related Works}

Many efforts have been made for word sense learning.
In terms of resources, several methods for multiple senses representation learning automatically induce word senses from monolingual corpora~\cite{reisinger2010multi,di2013clustering}. 
\cite{reisinger2010multi} provided a context-dependent vector representation of word meaning with the Wikipedia and Gigaword corpus.\cite{di2013clustering} uses the automatic discovery of word senses to Web search result clustering.
Others extended it to using bilingual corpora~\cite{guo2014learning, neelakantan2014efficient, ettinger2016retrofitting, vsuster2016bilingual,kartsaklis-etal-2018-mapping,camacho2018word}.
\cite{guo2014learning} proposed to learn sense-specific word embeddings by exploiting bilingual resources.
\cite{neelakantan2014efficient} presented an extension to the Skip-gram model to learn multiple embeddings per word type.
\cite{ettinger2016retrofitting} proposed to retrofit sense-specific word vectors using parallel text.
\cite{vsuster2016bilingual} used bilingual learning of multi-sense embeddings with discrete autoencoders.
These methods focus on the statistics information extracted from text corpora and contribute to the development of the word sense learning research area greatly. However, at the same time, they ignore exploring knowledge from semantic networks~\cite{mancini2017embedding}. As a result, the induced senses are not readily interpretable and are not easily mappable to lexical resources either~\cite{panchenko2017unsupervised}.

By taking advantage of the information in a knowledge base, the knowledge-based approaches are proposed to exploit knowledge resources like WordNet and BabelNet~\cite{agirre2014random, moro2015semeval, mancini2017embedding, pasini2018two,kumar-etal-2019-zero,schwab2019sense,wang2020word}.
\cite{agirre2014random} presented a WSD algorithm based on random walks over large Lexical Knowledge Bases (LKB).
\cite{moro2015semeval} analyzed how using a resource that integrates BabelNet might enable WSD to be solved.
\cite{pasini2018two} presented two fully automatic and language-independent sense computing methods based on BabelNet and Wikipedia.
\cite{mancini2017embedding} exploited large corpora and knowledge from the semantic networks to produce word and sense embeddings. 
The multi-embeddings based methods usually rely on the knowledge base when learning multiple embeddings
for each sense~\cite{jauhar2015ontologically, pelevina2016making, reisinger2010multi}.
The advantage of the knowledge-based systems is that they do not require sense-annotated data, and the knowledge enhances the word sense learning ability of these methods. 
However, without the sense-annotated data in these methods, the performance is also limited, and each disambiguation word is treated in isolation with a weak relationship~\cite{raganato2017neural}).

In order to address deficiencies of the knowledge-based approaches, some works exploiting the use of the sense annotated corpus have been proposed. In works that use supervised data, a machine learning
classifier is trained with a large amount of data with annotated senses~\cite{kartsaklis-etal-2013-separating, yuan2016semi, raganato2017neural}.
Usually, they depend greatly on the annotated corpus, e.g. SemCor3.0 and OMSTI. It is at the expense, however, of harder training and limited flexibility~\cite{liu-etal-2018-lcqmc}.

As mentioned above, the monolingual and bilingual corpora, the knowledge base, and the sense annotated corpus improve the word sense learning ability of the methods. 
Combining these resources can further improve the performance of sense learning methods. For example, even though the BERT model~\cite{devlin2019bert} shows outstanding performance on word sense learning due to the language pre-training, even better performance can be reached when the knowledge base or supervised corpus is integrated into the BERT model~\cite{Daniel2019Language, HuangGlossBERT, scarlini2020sensembert}. \cite{Daniel2019Language} focused on the synset, hypernym, and lexname with full-coverage of WordNet on BERT. ~\cite{HuangGlossBERT} focused on better leveraging gloss knowledge into the BERT model and used the SemCor3.0 corpus to train the model. Both methods have provided the new state-of-the-art on the word sense learning task in succession.

All the above methods have made a great contribution to the process of word sense learning. 
Our work is highly related to the previous ones but we propose a word sense learning method from a new perspective. In our proposed method, we learn the appropriate sense of an ambiguous word from the unsupervised word embedding with the powerful capsule network and contextual attention, and the method is trained with the sense-annotated corpus by word sense matching training. We have found that using unsupervised word embedding with the capsule network can reach the new state-of-the-art on word sense learning tasks. However, we are unaware of any such methods in word sense learning.

\section{Method}
\begin{figure*}
 \centering
 \includegraphics[width=0.9\textwidth]{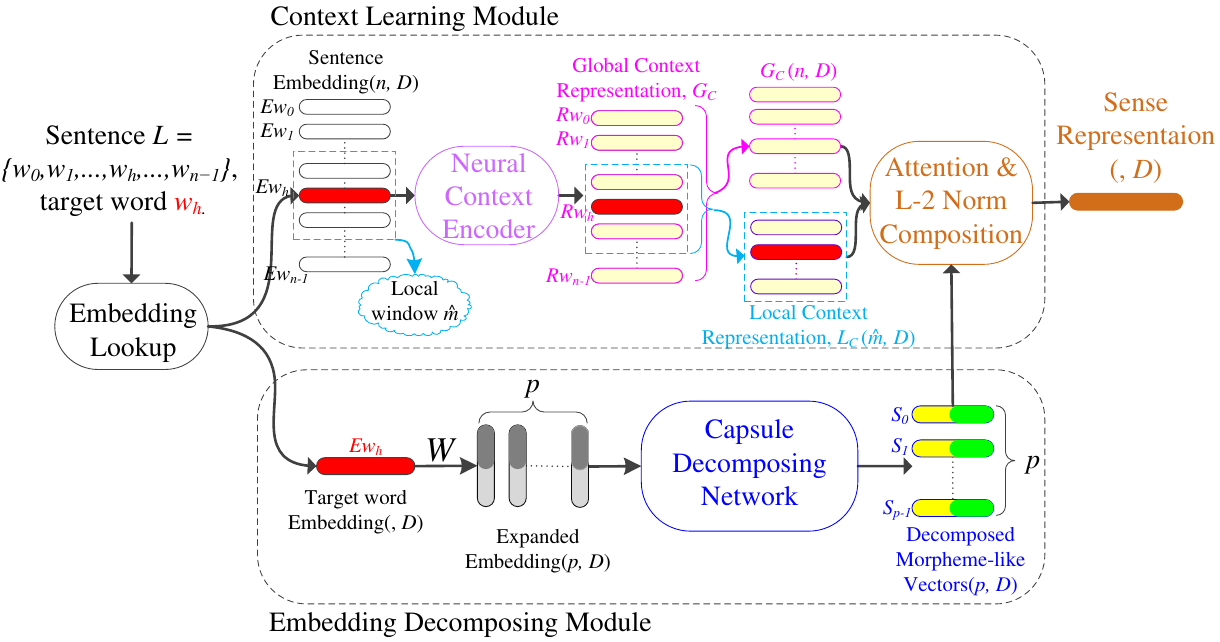}
 \caption{\label{DecE2S}The embedding decomposing and context learning procedure of the CapsDecE2S model. The numbers in the bracket mean the variable dimension.}
\end{figure*}

Figure~\ref{DecE2S} depicts an overall diagram of the proposed CapsDecE2S method.
There are two main modules in the CapsDecE2S method: the embedding decomposing and the context learning.

In CapsDecE2S, the ambiguous word in the sentence is regarded as the target word. With the embedding lookup, we can derive each word embedding in the sentence.
The target word embedding is fed into the embedding decomposing module and 
to the context learning module.
The embedding decomposing module decomposes the target word embedding with the capsule network and outputs the multiple morpheme-like vectors.
At the same time, in the context learning module, each word embedding in the sentence is used to first derive the global and local context representation, and then together with the decomposed morpheme-like vectors from the decomposition module to learn the context-specific sense representation of the target word.

In Section 3.1, we describe how to decompose the unsupervised embedding with the capsule network in detail and in section 3.2, we describe the context learning procedure. In Section 3.3, we introduce the proposed word sense matching training on how to train the CapsDecE2S method.

\begin{figure}
 \centering
 \includegraphics[width=0.7\textwidth]{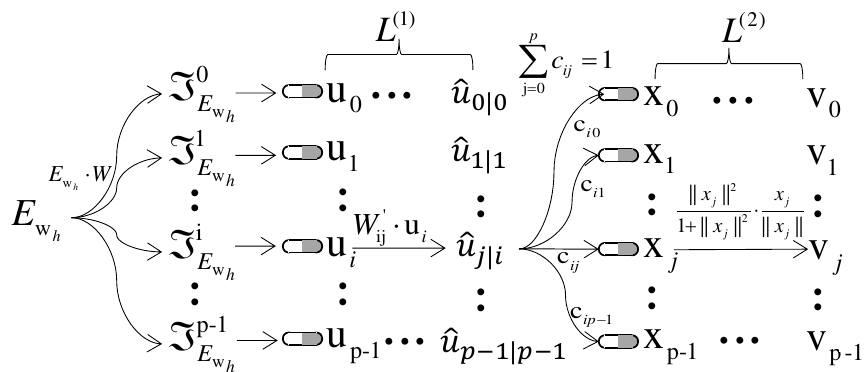}
 \caption{\label{capsule_qq}The decomposing calculation between capsules in the initial two layers. In the following layer, the outputs $\{v_0, v_1,...,v_{p-1}\}$ are used as its input, namely $u_k = v_k, k \in [0,p) $.}
\end{figure}

\subsection{The Embedding Decomposing with the Capsule Network}
In the embedding decomposing module shown in Figure~\ref{DecE2S}, the target word embedding is first expanded by parameters $W$. Next, the expanded embeddings are input into the capsule decomposing network. Then, the capsule decomposing network outputs the decomposed morpheme-like vectors.

Figure~\ref{capsule_qq} depicts the decomposing calculation of two initial layers in the capsule decomposing network.
$L = \{w_0, w_1,$ $..., w_h, ..., w_{n-1}\}$ denotes a sentence, and the target word $w_h$ embedding is $E_{w_h}$.
First, $E_{w_h}$ is expanded with parameter $W$ as $\Im_{E_{w_h}}^{i}=E_{w_h}\cdot W_i, i\in [0,p)$, where $p$ is the maximum number of the decomposed vectors.
In the first layer, the input is $\Im_{E_{w_h}}$, and each $\Im_{E_{w_h}}^{i}$ corresponds to one capsule.
For a capsule $i$ in the layer $L^{(1)}$(abbr. $i_{L^{(1)}}$), we have $u_i = \Im_{E_{w_h}}^{i}$.
A weight matrix $W_{ij}^{'}$ is used for building connections with  the capsule $j$ in the layer $L^{(2)}$(abbr. $j_{L^{(2)}}$), and a prediction vector $\hat{u}_{j|i}$ is produced.
Next, the total input $x_j$ to the capsule $j_{L^{(2)}}$ is a weighted sum over all $\hat{u}_{j|i}$ from the capsules in the layer $L^{(1)}$.

\begin{equation} \label{eqng}
 x_j = \sum_{i}c_{ij}\hat{u}_{j|i},\quad \hat{u}_{j|i} = W_{ij}^{'}u_i,
\end{equation}
where $c_{ij}$ is the coupling coefficient from capsule $i_{L^{(1)}}$ to $j_{L^{(2)}}$. The coupling coefficients sum to 1 between $i_{L^{(1)}}$ and all capsules in $L^{(2)}$.

\begin{equation} \label{eqncij}
\sum_{j=0}^{p}{c_{ij}} = 1.
\end{equation}

In the capsule $j_{L^{(2)}}$, a non-linear "squashing" function is applied to keep the length by shrinking short vectors to almost 0 and long vectors to a length slightly below 1, is shown in Equation~\ref{eqngvf}.
\begin{equation} \label{eqngvf}
 v_j = \frac{\lVert{x_j}\rVert^2}{1+\lVert{x_j}\rVert^2}\cdot\frac{x_j}{\lVert{x_j}\rVert},
\end{equation}
where $v_j$ is the squashing output of the capsule $j_{L^{(2)}}$.

The coupling coefficient $c_{ij}$ is updated by the iterative dynamic routing, and it is a softmax result based on the logic $b_{ij}$.
\begin{equation} \label{eqng}
 c_{ij} = \frac{exp(b_{ij})}{\sum_{k=0}^{p}{exp(b_{ik})}},
\end{equation}
we follow the processing by~\cite{sabour2017dynamic}. Initially, $b_{ij}$ equals to 0 and is updated as

\begin{equation} \label{eqng}
b_{ij} = b_{ij} + v_j\cdot \hat{u}_{j|i},
\end{equation}
which aims to measure the agreement between the output $v_j$ of $j_{L^{(2)}}$ and the prediction $\hat{u}_{j|i}$ of $i_{L^{(1)}}$.

In the following layers, the network repeats the same calculation. The output $v$ is passed into the capsules in the next layer
and goes through the weight matrix, the weighted sum and the non-linear squashing function.
With $K$ layer iterations, we take the outputs of layer $K$ as the decomposed vectors $\{S_0,S_1,...,S_{p-1}\}$, where $S_j = v_{j_{L^{(K)}}}$.

\subsection{The Context Learning Module}
In the context learning module shown in Figure~\ref{DecE2S}, the word embeddings in the sentence are input into the neural context encoder. The output of the neural context encoder is used as the global context representation, and the nearby word representations of the target word are used as the local context representation.
Finally, the decomposed morpheme-like vectors, the global context representation, and the local context representation are input into the "Attention\&L-2 Norm Compostition", which then outputs the sense representation.

Here, we introduce the calculations in the context learning module in detail.
We take the neural language model (NLM) as the neural context encoder to learn the context information.

First, the words in a sentence $L$ are converted into $\{E_{w_0}, ...,$ $E_{w_{n-1}}\}$ by embedding lookup, and then passed into the neural context encoder seriatim.

Second, the hidden states of the neural units in the last layer of the neural context encoder are selected as the context representation. Here, all the word representations are regarded as the
global context representation $G_c$, namely $G_c = \{R_{w_0},...,R_{w_{n-1}}\}$.

Third, we extract the nearby word representations of the target word $w_h$ as the local context representation $L_c$ with a window size $\hat{\mathfrak{m}}$, namely $L_c = \{R_{w_e},...,R_{w_h},...,R_{w_z}\}$, where $e = min(h-\hat{\mathfrak{m}}/2,0), z = max(h+\hat{\mathfrak{m}}/2,n)$.

Forth, the "Attention\&L-2 Norm Composition" component first calculates the global context attention weight and the local context attention weight on the decomposed morpheme-like vectors $\{S_0,$ $S_1,...,S_{p-1}\}$ based on the global context representation $G_c$ and the local context representation $L_c$.
The global attention weight $a^{G} = \{a_0^{G},...,a_k^{G},..,a_{p-1}^{G}\}$ is calculated as
\begin{equation} \label{global}
 a_k^{G} = \frac{exp({\hat{c}}_{k})}{\sum_{j=0}^{n}exp({\hat{c}}_{k_j})}, \quad {\hat{c}}_{k_j} = {S_k}\cdot{G_{c_j}},
\end{equation}
where $S_k$ is one decomposed morpheme-like vector in $\{S_0,$ $S_1,...,S_{p-1}\}$ and ${G_{c_j}}$ is the $j-th$ representation in the global context representation $G_c$.
The local attention weight $a^{L}$ =  $\{a_0^{L}$, ..., $a_k^{L}$,.., $a_{p-1}^{L}\}$ is also calculated in a similar way as
\begin{equation} \label{local}
 a_k^{L} = \frac{exp({\hat{c}}_{k}^{'})}{\sum_{j=0}^{n}exp({\hat{c}}_{k_j}^{'})}, \quad {\hat{c}}_{k_j}^{'} = {S_k}\cdot{L_{c_j}},
\end{equation}
where ${L_{c_j}}$ is the $j-th$ representation in the local context representation $L_c$.

Next, we apply the attention weights $a^{G}$ and $a^{L}$ to its global and local context representations respectively, and obtain the context-specific vectors $\mathcal{S}^{*}$ as
\begin{equation} \label{eqng}
 \mathcal{S}_k^{*} = S_k + \sum_{i=0}^{n}a_{k_i}^{G}\cdot G_{c_i} + \sum_{i=0}^{z-e+1}a_{k_i}^{L}\cdot L_{c_i}.
\end{equation}

Finally, we use the L-2 norm of each $\mathcal{S}^{*}$ to represent the weight $\hat{b} $= $\{\hat{b}_0$, ..., $\hat{b}_k$, ..., $\hat{b}_{p-1}\}$ in composing the final sense representation $Q_\mathcal{S}$ in its context.
\begin{equation} \label{eqng6}
\begin{split}
  Q_\mathcal{S} = \sum_{k=0}^{p}\hat{b}_k\cdot\mathcal{S}_k^{*},\hat{b} = \frac{exp(\hat{b})}{\sum_{k=0}^{p}exp(\hat{b}_{k})},
  \hat{b}_k = \lVert\mathcal{S}_k^{*}\rVert^2.
  \end{split}
\end{equation}

\subsection{Word Sense Matching Training}

\begin{figure}
 \centering
 \includegraphics[width=0.6\textwidth]{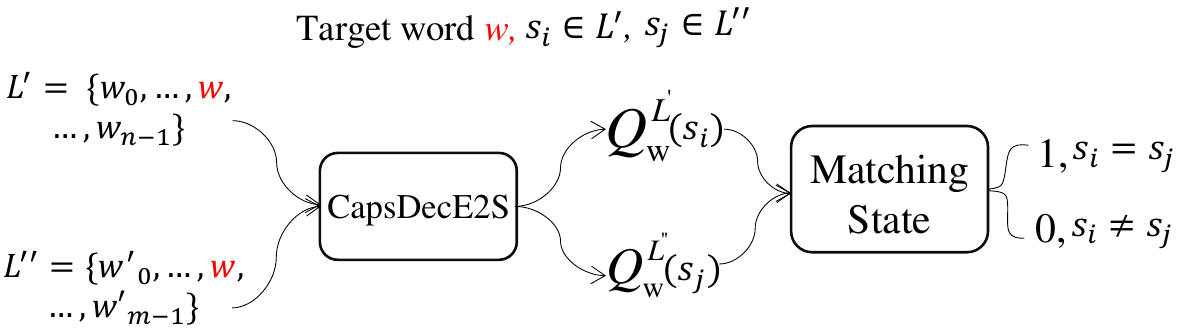}
 \caption{\label{WordMatching}The word sense matching training process to predict the matching state of the word senses for $w$ in $L'$ and $L''$, respectively. The candidate senses of word $w$ is $\{s_0, s_1, ..., s_m\}$ and its true senses in $L'$ and $L''$ are $s_i$ and $s_j$.}
\end{figure}

Figure~\ref{WordMatching} depicts the word sense matching training process to learn the context-specific sense representation.
Two sentences that contain the target word are input the CapsDecE2S model respectively, and CapsDecE2S outputs the sense representation for each target word, which is the output of the "Attenion\&L-2 Norm Composition" in context learning module.
The procedure is the same as in Figure~\ref{DecE2S}.
Two sense representations are used to train the model by predicting their matching state. More details on the word sense matching training are given below.

In Figure~\ref{WordMatching}, the sentences, e.g. $L^{'}$ and $L^{''}$, usually consist of a series of words. The word $w$ is the target ambiguous word that we need to disambiguate.
For $w$, its candidate sense set is $Set(w) = \{s_0, s_1, ..., s_m\}$, which are defined in WordNet. The true word senses $s_i$ and $s_j$ in both sentences satisfy $s_i\in L^{'}$ and $s_j\in L^{''}$.

In word sense matching training, we define the matching and non-matching state $M$ as:
\begin{equation} \label{eqng}
  M =
\begin{cases}
1,  & \mbox{if }s_i = s_j \\
0, & \mbox{if }s_i \neq s_j
\end{cases}
\end{equation}

The matching prediction, based on the sense representations $Q_{\mathcal{S}}$ in $L^{'}$ and $L^{''}$, is used for the cross-entropy loss.
First, the learned word sense representations in $L^{'}$ and $L^{''}$ are represented as $Q_w^{L^{'}}$ and $Q_w^{L^{''}}$, respectively. 
Second, we concatenate $Q_w^{L^{'}}$ and $Q_w^{L^{''}}$, and pass it into the softmax classifier.
A binary probability $\hat y$ is predicted.
Finally, based on the matching state $M$, the gold label is converted into binary $y$, and the cross-entropy loss is
\begin{equation}\label{generator}
\mathcal{L} = -[ylog\hat y+ (1-y)log(1-\hat y)]
\end{equation}

\section{Experiments and Results}

\subsection{Datasets and Setup}

We evaluated the proposed CapsDecE2S model on~\cite{pilehvar2018wic}'s Word-in-Context (WiC) dataset with the metric of accuracy and~\cite{raganato2017word}'s cross-domain English all-words Word Sense Disambiguation (WSD) datasets
(Senseval-2 (SE2)~\cite{semeval-2001-senseval},
Senseval-3 task1 (SE3)~\cite{ws-2004-senseval},
SemEval-07 task 17 (SE07)~\cite{pradhan-etal-2007-semeval},
SemEval-13 task12 (SE13)~\cite{navigli-etal-2013-semeval},
and SemEval-15 task13 (SE15)~\cite{moro2015semeval}) with the metric of F1 score.

The WiC dataset~\cite{pilehvar2018wic} is a new benchmark for the evaluation of context-sensitive word embeddings (Train:5.4K, Dev:0.63K, Test:1.4K), the target words in the dataset are nouns and verbs. On WiC, we used the default training data and tested online~\footnote{https://competitions.codalab.org/competitions/20010}.
We compare our results to the methods cited from the relevant papers~\cite{chang-chen-2019-word,Daniel2019Language,pilehvar2018wic}.

On WSD datasets, following most researches~\cite{HuangGlossBERT,Daniel2019Language,luo2018incorporating,raganato2017word}, we used the SemCor3.0 corpus as the training set and SE07 as dev set. The SemCor3.0 corpus is the largest manually annotated corpus with WordNet sense for WSD, widely used by~\cite{ iacobacci2016embeddings, luo-etal-2018-leveraging, luo2018incorporating, raganato2017word, raganato2017neural, zhong2010makes}.
The published methods~\cite{iacobacci2016embeddings,luo-etal-2018-leveraging,raganato2017neural} and SOTA~\cite{Daniel2019Language} are used for comparison.

Hyper-parameters in the experiments are as follows: the number $p$ of the decomposed vectors (10); The capsule network layers (3); The routing iterations in capsule network (3); Unless specified otherwise, the pre-trained uncased BERT base and large are used
as the neural context encoder, respectively; Especially, in WordPiece tokenization, we use the head of the sub-token as the target word. As for the others, we follow the default settings.

\subsection{Baselines and Comparison Methods}

The CapsDecE2S model was competed among its different versions which are implemented in this paper, along with other published methods.
These published methods include the supervised models for classification, e.g., {{{GlossAug.S.}}{ {(GAS)}}}~\cite{luo2018incorporating},  {{{Hier.Co-Att.}}{{(HCAN)}}}~\cite{luo-etal-2018-leveraging} and $\mbox{EWISE}_{{ConvE}}$~\cite{kumar-etal-2019-zero}, the supervised models for learning contextual representations of senses, e.g., Context2vec~\cite{Melamud2016context2vec}, ELMo~\cite{peters2018deep}, BERT~\cite{devlin2019bert} and {GLU(1sent+1sur)}~\cite{hadiwinoto-etal-2019-improved}, and the aggregation models with knowledge base, e.g., ${\mbox{{LMMS}}}_{2348}$~\cite{Daniel2019Language}, SemCor + hypernyms~\cite{schwab2019sense}, and GlossBERT~\cite{HuangGlossBERT}. 
All these methods are under the same evaluation protocol as~\cite{pilehvar2018wic}, and~\cite{raganato2017word} on both tasks.
The methods that used the additional manually-annotated corpus except SemCor3.0 for training and the ensemble versions are not included, e.g. partial methods in~\cite{schwab2019sense} and~\cite{scarlini2020sensembert}, but not limited to those that use the knowledge base and unsupervised resources.
The proposed CapsDecE2S method and its different versions that are implemented in this paper are described as below.

\noindent$\bullet $\quad {{${CapsDecE2S_{base}}$:} 
This is our proposed model that all components are introduced in Section "Method", and the neural context encoder that here we used for context learning is the BERT base.
The parameters for BERT base are provided by the following official link\footnote{https://storage.googleapis.com/bert\_models/2018\_10\_18/ uncased\_L-12\_H-768\_A-12.zip}.}

\noindent$\bullet $\quad  {{${CapsDecE2S}_{large}$:}}
The main difference from CapsDec-
$E2S_{base}$ is that in this model the neural context encoder for context learning is the BERT large. 
The parameters for BERT large are provided by the following official link\footnote{https://storage.googleapis.com/bert\_models/2018\_10\_18/uncased\_L-24\_H-1024\_A-16.zip}.

\noindent $\bullet$ \quad  \mbox{$CapsDecE2S_{base}+LMMS$}:
LMMS is the sense-level embeddings with full-coverage of WordNet~\cite{Daniel2019Language} and can be obtained from the link~\footnote{https://github.com/danlou/lmms}. In LMMS, each sense is allocated an embedding. In this model, the LMMS embedding is concatenated with the corresponding target word sense representation, $Q_{w^{L^{'}}}$ or $Q_{w^{L^{''}}}$ in figure~\ref{WordMatching}, in both sentences before they are passed into the softmax classfication. First, we compute the similarity between the corresponding sentence and the LMMS sense embedding under the same lemmas, and then the most similar LMMS sense embedding is selected for concatenation. The other settings are the same as the ${CapsDecE2S_{base}}$.

\noindent$\bullet $\quad {$\mbox{CapsDecE2S}_{large}$+\mbox{LMMS}:}
The difference from the Caps
-$DecE2S_{base}$+LMMS is that in this model the neural context encoder for context learning is the BERT large.

\noindent$\bullet $\quad  {$\mbox {BERT}_{base}^{wsm}$:} The BERT base model is fine-tuned with the word sense matching training(wsm), which the sentence pair goes through the BERT base model. Then we extract the target word representations in both sentences and concatenate them as input to the softmax classification for matching training.

\noindent$\bullet $\quad  {$\mbox {BERT}_{large}^{wsm}$:} 
The model is trained in the same way as the {${BERT}_{base}^{wsm}$} model, but it is BERT large model.

\begin{table}[t!]
\begin{center}
\begin{tabular}{p{0.1cm}lc}
\hline &\bf  Method& \bf  Accuracy\\ \hline
\multirow{2}{*}{\small \rotatebox{90}{S-Lv} }&$\mbox {\underline{1.BoW}}^{\dag}$&  $58.7^{\dag}$\\
&$\mbox {\underline{2.Sentence LSTM}}^{\dag}$&  $54.0^{\dag}$\\
\hline
\multirow{3}{*}{\small \rotatebox{90}{Multi-E} }&$\mbox {\uwave {3.DeConf}}^{\dag}$~\cite{pilehvar2016conflated}&  $58.7^{\dag}$\\
&$\mbox {\uwave {4.SW2V}}^{\dag}$~\cite{mancini2017embedding}&  $58.1^{\dag}$\\
&$\mbox {\uwave {5.JBT}}^{\dag}$~\cite{pelevina2016making}&  $54.7^{\dag}$\\
\hline
\multirow{6}{*}{\small \rotatebox{90}{Contextual}}&$\mbox {\dashuline {6.Context2vec}}^{\dag}$~\cite{Melamud2016context2vec}&$59.3^{\dag}$\\
&$\mbox {\dashuline {7.ELMo}}_{{3/1}}^{\dag}$~\cite{peters2018deep}&  $58.0/57.0^{\dag}$\\
&$\mbox {\dashuline {8.BERT}}_{{base/large}}^{\dag}$~\cite{devlin2019bert}&  $65.4/65.5^{\dag}$\\
&$\mbox {\dashuline {9.TextCNNBERT}}$~\cite{chang-chen-2019-word}&  $68.6$\\
&$\mbox {\dashuline {10.LMMS}}_{{2348}}^{*}$ ~\cite{Daniel2019Language}&  $69.1^{*}$\\
\hline
&{$\mbox{11.CapsDecE2S}_{base}$} &\bf 70.6\\
\hline
\end{tabular}
\end{center}
\caption{\label{wic_results}\small{Comparisons of accuracy (\%) on the WiC dataset. \dag - cited from WiC paper~\cite{pilehvar2018wic} with the best results and others from the corresponding papers. * - the authors only reported the dev result. The methods with an underline, wavy-line, and dash-line correspond to the sentence-level baselines (S-Lv), multi-embedding based models (Multi-E), and contextualized word based models (Contextual), respectively.
}}
\end{table}

\subsection{Results on Word in Context}
Table~\ref{wic_results} lists the accuracies of the sentence-level baselines, multi-embedding based models, contextualized word based models, and our {$\mbox{CapsDecE2S}_{base}$} model.
Since there are formal training corpus with 5.4K pairs on WiC task, we only report the {$\mbox{CapsDecE2S}_{base}$} model without using any other resources. 
{$\mbox{CapsDecE2S}_{base}$} outperforms all the other methods by a large margin with an accuracy of 70.6\% (the {$\mbox{CapsDecE2S}_{large}$} model performs quite close to this score in this task, so we do not report it.).

The sentence-level models are widely used for sentence encoding, but they show poor performance. The main reason may be that the WiC dataset is too small, and the scale limits such methods without any pre-training.
The multi-embedding based models make use of external lexical resources, which helps to learn more accurate sense.
The contextualized word based models benefit from the large-scale language pre-training, and thus show better performance than methods foregoing.
Especially, the method $\mbox { {LMMS}}_{{2348}}$ based on BERT large uses additional information from Wordnet, the dictionary embedding, and the fastText embedding.
The main difference between {$\mbox{CapsDecE2S}_{base}$} and BERT-based models, e.g. $\mbox {{TextCNNBERT}}$ and $\mbox {{LMMS}}_{{2348}}$ (BERT), is the capsule decomposing module.
{$\mbox{CapsDecE2S}_{base}$} contributes to the WiC dataset with more than 1.5\% absolute improvement in the accuracy.
The result indicates the information contained in the unsupervised word embedding has a potential
under limited training data and the learning ability of the capsule network.

\begin{table}
\begin{center}
\begin{tabular}{p{0.2cm}|l|c|cccc|c}
\hline &\bf\small  Method &\bf\small SE07& \bf \small SE2& \bf \small SE3&\bf\small SE13&\bf \small SE15&\bf \small ALL\\ \hline

&\small {\mbox{ {1.MFS baseline}}}&\small 54.5&\small 65.6& \small 66.0& \small 63.8& \small 67.1& \small 64.8\\
\hline\hline

\multirow{6}{*}{\small \rotatebox{90}{$Sup$} }&\small {{ 2.{IMS+emb}}}\small~\cite{iacobacci2016embeddings}&\small 62.6&\small 72.2& \small 70.4&\small  65.9& \small 71.5&\small  69.6\\

&$\small {\mbox{ 3.{Seq2Seq}}}_{multi-tasks}$ \small~\cite{raganato2017neural}&\small63.1 & \small 70.1& \small 68.5&\small  66.5& \small 69.2& \small  68.6\\

&$\small {\mbox{ 4.{Bi-LSTM}}}_{multi-tasks}$\small~\cite{raganato2017neural}&\small 64.8& \small 72.0& \small 69.1&\small  66.9&\small  71.5& \small 69.9\\

&\small {\small { 5.{GlossAug.S.}}{\small {(GAS)}}}\small~\cite{luo2018incorporating}&\small -& \small 72.2& \small 70.5&\small 67.2& \small 72.6& 70.6\\

&\small {\small{ 6.{Hier.Co-Att.}}{\small {(HCAN)}}}\small~\cite{luo-etal-2018-leveraging}&\small -&\small  72.8& \small 70.3&\small  68.5& \small 72.8& 71.1\\

&\small {{$\mbox{7.EWISE}_{{ConvE}}$}}~\cite{kumar-etal-2019-zero}&\small 67.3&\small 73.8&\small 71.1&\small 69.4&\small 74.5&\small 71.8\\
\hline\hline


\multirow{3}{*}{\small \rotatebox{90}{$Sup_{c}$} }&\small ${\mbox{ 8.{BERT}}}_{Token-CLS}$~\cite{HuangGlossBERT}&\small 61.1&\small 69.7&\small 69.4&\small 65.8&\small 69.5&\small 68.6\\

&\small {{ {9.Context2Vec}}}\small~\cite{Melamud2016context2vec}&\small 61.3& \small 71.8& \small 69.1&\small 65.6& \small 71.9& \small 69.0\\

&\small {10.GLU(1sent+1sur)}~\cite{hadiwinoto-etal-2019-improved}&\small 68.1& \small 75.5&\small  73.6&\small  71.1&\small 76.2& \small 74.1\\

\hline\hline

\multirow{5}{*}{\small \rotatebox{90}{$Sup_{c}+KB$} }&\small ${\mbox{ 11.{ELMo k-NN}}}_{full-cover.}$\small~\cite{Daniel2019Language}&\small 57.1&\small  71.5&\small  67.5& \small 65.3& \small 69.6& \small 67.9\\

&\small ${\mbox{{12.BERT k-NN}}}_{full-cover.}$\small~\cite{Daniel2019Language}&\small 66.2& \small 76.3&\small  73.2&\small  71.7&\small  74.1& \small 73.5\\

&\small ${\mbox{{13.LMMS}}}_{2348}$\small~\cite{Daniel2019Language}&\small 68.1& \small 76.3&\small  75.6&\small  75.1&\small  77.0& \small 75.4\\

&\small {{{14.SemCor+hypernyms}}}~\cite{schwab2019sense}&\small 69.5& \small 77.5&\small  77.4&\small  76.0&\small  78.3& \small 76.7\\

&\small {15.GlossBERT}~\cite{HuangGlossBERT}&\small 72.5&\small 77.7&\small 75.2&\small 76.1&\small \bf 80.4&\small 77.0\\

\hline\hline
\multirow{6}{*}{\small \rotatebox{90}{$Ours$} }&\small $\mbox {16.BERT}_{base}^{wsm}$&\small 59.2&\small 73.3&\small 72.2&\small 65.3&\small 72.6&\small 69.8\\

&\small $\mbox {17.BERT}_{large}^{wsm}$&\small 59.6&\small 73.9&\small 72.4&\small 65.3&\small 73.1&\small 70.6\\

&\small {$\mbox{18.CapsDecE2S}_{base}$}&\small 67.0&\small 77.4&\small 76.2&\small  75.9& \small 77.3&\small 76.1\\

&\small {$\mbox{19.CapsDecE2S}_{large}$}&\small 68.7&\small \bf 78.9& \small 77.4&\small  75.6&\small 77.1&\small 76.9\\

&\small {$\mbox{20.CapsDecE2S}_{base}$+LMMS}&\small 73.5&\small78.4& \small 79.4&\small  76.5&\small78.6&\small 77.9\\

&\small {$\mbox{21.CapsDecE2S}_{large}$+LMMS}&\small \bf 73.8&\small  78.8& \small \bf80.7&\small\bf 76.6&\small 79.4&\small \bf 78.6\\
\hline
\end{tabular}
\end{center}
\caption{\label{wsd}\small{Comparison in terms of F1-score(\%) on the English all-words WSD test sets of~\cite{raganato2017word} under the manually-annotated training corpus SemCor3.0. Methods are grouped by the types of, supervised models for classification ($Sup$), supervised models for learning contextual representations of senses ($Sup_{c}$), the aggregation models of $Sup_{c}$ and knowledge base ($Sup_{c}+KB$), and our implementations~\cite{scarlini2020sensembert}. For each method, we list the version with best results.
}}
\end{table}

\subsection{Results on English all-words WSD}

Table~\ref{wsd} lists the comparison results in terms of F1-score on the English all-words WSD test sets between different types of methods published in recent years.
As we can see, the CapsDecE2S based models show promising results on each test set. The {$\mbox{CapsDecE2S}_{large}$+LMMS} reaches the best results on nearly all test sets when compared to other approaches, and outperforms the state-of-the-art (Line 15) with 1.1\% improvement on "SE2", 5.5\% on "SE3", 0.5\% on "SE13", and 1.6\% on "ALL".

Furthermore, most "$Sup_c$" based models (Line 8-15) outperform the "$Sup$" models (Line 2-7), which proves the ability of the contextual representation of senses. However, the single BERT model shows poor performance (Line 8,16,17) no matter it is trained by context-gloss training~\cite{HuangGlossBERT}, k-NN training~\cite{Daniel2019Language}, or word sense matching training. The GLU (1sent+1sur) model explores multiple strategies of integrating BERT contextualized word representations and reaches better results~\cite{hadiwinoto-etal-2019-improved}.
When the information of the knowledge base, e.g., WordNet, is integrated into the BERT model (Line 13-15), all these models show outstanding performances. For example, the GlossBERT model leverages
the gloss knowledge from WordNet in a supervised neural WSD system~\cite{HuangGlossBERT}. The ${\mbox{{LMMS}}}_{2348}$ model focuses on creating sense-level embeddings with full-coverage of WordNet~\cite{Daniel2019Language}. The SemCor+hypernyms model exploits the semantic
relationships, e.g., synonymy, hypernymy and hyponymy, to compress the sense vocabulary of WordNet~\cite{schwab2019sense}.

The CapsDecE2S model is a supervised model of learning contextual representations of senses, and it explores the information from the unsupervised word embedding by the capsule network. Obviously, in this way, the model outperforms the "$Sup_c$" and "$Sup$" models and shows competitive results with the "$Sup_{c}+KB$" models.
When the knowledge base embedding is integrated into the CapsDecE2S model, named {$\mbox{CapsDecE2S}$+LMMS} (Line 20,21), the new state-of-the-art are reached on "SE2", "SE3". "SE13" and "ALL" sets. The result on "SE15" is also encouraging.


\section{Discussion}

In this section, we perform analyses and schematize several examples to quantitatively interpret some properties of the CapsDecE2S model, 
including the ablation study on the CapsDecE2S' components, 
the CapsDecE2S' sense similarity across contexts, 
the attention weight in context,  
the cases that CapsDecE2S fails to learn.

\subsection{Ablation Study on the CapsDecE2S' Components}
First, we perform an ablation study on the CapsDecE2S' components on the Enghlish "ALL" test set in~\cite{raganato2017word} to see how each of the components affects the final F1-score in Table~\ref{ablation}.
Experiments are conducted based on $\mbox{CapsDecE2S}_{base}$ by removing the global context, the local context, the capsule decomposing module, and the word sense matching training, respectively.
The "w/o capsule network" means that the expanded embeddings are directly passed into the "Attention\&L-2 Norm Composition" without capsule decomposing in Figure~\ref{DecE2S}. 
The "w/o Word Sense Matching" means the capsDecE2S method was not trained by word sense training but by the sense classification training. 
The last column "Abs/Rel (\%)" in Table~\ref{ablation} gives the absolute and relative improvement, that are commonly used to measure the change or difference of two variables, on the F1 score compared to $\mbox{CapsDecE2S}_{base}$. 

From Table~\ref{ablation}, we can see that each component plays an important role in $\mbox{CapsDecE2S}_{base}$. Without the capsule network, the model drops greatly with 6.2\% deterioration, and the result is quite close to $\mbox {BERT}_{base}^{wsm}$ in Table~\ref{wsd}, which proves the capsule decomposing ability. 
The local context seems to be more useful than the global context with F1 score of 73.6\% in "w/o Global context" and 72.8\% in "w/o Local context", which may be that the  texts far away from the target word matter less but exist as noise in the global context. Besides, one can also see that the method with sense classification training shows worse performance than that is with the word sense matching training.

\begin{table}[t!]
\begin{center}
\begin{tabular}{lcc}
\hline \bf  \small Method & \bf\small  F1&\small{\bf{Abs/Rel(\%)}}\\ \hline
\small $\mbox{CapsDecE2S}_{base}$ &\small 76.1&\small 0.0 /0.0\\
 \hline
 \small \quad w/o Global Context&\small 73.6&\small -2.5/-3.3\\
 \small \quad w/o Word Sense Matching&\small 72.9&\small -3.2/-4.2\\
 \small \quad w/o Local Context&\small 72.8&\small -3.3/-4.3\\
 \small \quad w/o Capsule Network&\small 69.9&\small -6.2/-8.1\\
\hline
\end{tabular}
\end{center}
\caption{\label{ablation}\small{Ablation study of CapsDecE2S’ components on
the "ALL" test set in terms of F1 score. "-" indicates the negative improvement.}}
\end{table}

\subsection{CapsDecE2S' Sense Similarity across Contexts}
Second, we conduct a sense similarity validation experiment to test the robust of the CapsDecE2S’ sense representation, in which we measure
the similarities of the CapsDecE2S’ sense representations when the sense occurs in different contexts. The experiment is based on the principle that even though one sense occurs in different sentences, its representations should hold high similarity.

In the experiment, we first randomly sampled twenty senses from "ALL" set, and each sense is allocated with at most five sentences.
Next, we randomly paired five sentences of each sense for three times and calculated the cosine scores between the target word representations.
Finally, we averaged the three cosine scores of each sense as its similarity value.
In the BERT model, the target word sense representation corresponds to the hidden embedding in the last layer.

\begin{figure}
 \centering
 \includegraphics[width=0.75\textwidth]{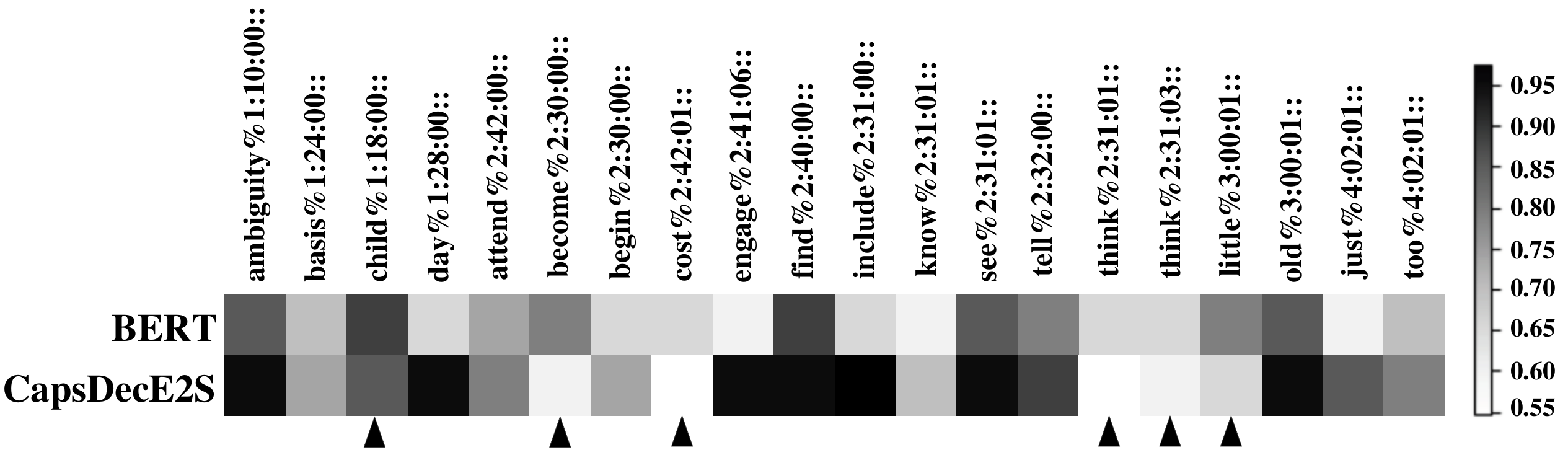}
 \caption{\label{cosine1}The visualized map of the sense similarity values from the BERT and CapsDecE2S models for sentences in "ALL" set. Column: the randomly selected senses. Color-bar: the sense similarity value scope.
}
\end{figure}

The visualized map of the sense similarity values calculated by BERT and CapsDecE2S is shown in Figure~\ref{cosine1}.
The format in the column to express the word sense is consistent with the definition in WordNet~\footnote{http://wordnetweb.princeton.edu/perl/webwn},
and the format explanation of each field can be found here~\footnote{https://wordnet.princeton.edu/documentation/senseidx5wn}.
For the contexts with more similar sense representations, their similarity value will be larger, and the color of the block in Figure~\ref{cosine1} will be darker.
From Figure~\ref{cosine1}, it is evident that apart from the 6 of 20 cases marked with a black triangle below, the remaining 16 of the blocks 
in the CapsDecE2S row are darker than the corresponding ones in the BERT model.
Besides, we could also see that the values by the CapsDecE2S model are usually located on the upper parts in the color-bar while those by the BERT model on the lower parts.
This experiment indicates that CapsDecE2S is more robust when learning the unique word sense.

\begin{figure}
\centering
\quad
\subfigure[S1: Probably ... "way": a line leading to a place or point]{
\begin{minipage}[t]{0.25\linewidth}
     \includegraphics[width=0.70\textwidth]{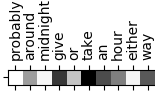}%
\end{minipage}
}
\quad
\subfigure[S2: I knew ... "way":  how something is done or how it happens]{
\includegraphics[width=0.45\textwidth]{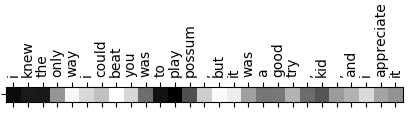}%
}
\caption{\label{way} The attentions weights on contexts given by s entences S1 and S2 for two senses of "way", respectively. The ellipsis "..." indicates the remainder of the contexts.}
\end{figure}

\subsection{Attention Weight in Context}
Third, we used a noun word "way" to analyze the relationship between its sense representation and the attentive words on the sense definition. 

Figure~\ref{way} schematizes the example word "way" and its two senses by analyzing the attention weights $a_k^{G}$ in Equation~\ref{global} on the context.
The titles of Figure~\ref{way}(a) and (b) give two sense definitions from the WordNet.
For either sense, we use the visualized map to present the attention weight distribution on the context when learning the context-specific sense representation.
The darker block means the attention weight value on this word is larger than the lighter ones.
The larger the value is, the more the context-specific sense representation relies on this word.

From Figure~\ref{way}(a) and (b), we can see that either sense relies on some words in the context, e.g. \{{\textit{"give", "take", "an", "hour"}}\} for sense (a) and \{{\textit{"i", "knew", "the", "only", "to", "play", "possum"}}\} for (b).
These words are essential in determining the unique semantic in the context, which proves that the context-specific sense representation indeed maintains the proper sense for its context.

\begin{table}[t!]
\begin{center}
\begin{tabular}{cll}
\hline
\bf\small  Word & \small\bf Sense definition&\small\bf Example Sentences\\ \hline

\multirow{5}{*}{\tiny Enough }&\multirow{2}{*}{\tiny {{Sufficient for the purpose}}}&{\tiny {{But there still are not {\emph {enough}} ringers to ring more than six}}}\\
&&{\tiny {{of the eight bells.}}}\\
\cline{2-3}
&\multirow{3}{*}{\tiny {{As much as necessary}}}&{\tiny {{Fortunately, these same parents do want their children to get }}}\\
&&{\tiny {{a decent education as traditionally understood, and they have }}}\\
&&{\tiny {{{\emph {enough}} common sense to know what that demands.}}}\\
\hline

\multirow{2}{*}{\tiny Shake }&{\tiny {{Move or cause to move back and}}}&{\tiny {{This time , it just got stronger and then the building started  }}}\\
&{\tiny {{forth}}}&{\tiny {{{\emph {shaking}} violently up and down.}}}\\
\cline{2-3}
&{\tiny {{Move with or as if with a tremor}}}&{\tiny {{My back is still in knots and my hands are still {\emph {shaking}}.}}}\\
\hline

\multirow{5}{*}{\tiny Plan }&{\tiny {{A series of steps to be carried out}}}&\multirow{2}{*}{\tiny {{U.N. group drafts {\emph {plan}} to reduce emissions}}}\\
&{\tiny {or goals to be accomplished}}&\\
\cline{2-3}
&{\tiny {{The act or process of drawing up}}}&{\tiny {{There were relatively few cases reported of attempts to involve }}}\\
&{\tiny {{plans or layouts for some project}}}&{\tiny {{users in service {\emph {planning}} but their involvement  in service }}}\\
&{\tiny {{or enterprise}}}&{\tiny {{provision was found to be more common.}}}\\
\hline

\end{tabular}
\end{center}
\caption{\label{close_senses} The words for which the CapsDecE2S failed to distinguish their close senses. Example sentences are from the WSD test sets and the target words are marked with the underline.}
\end{table}

\subsection{Cases that CapsDecE2S Fails to Learn}
Finally, our experiments and analyses have proven the sense learning ability of the CapsDecE2S model, but the experimental results also imply that CapsDecE2S is not omnipotent.
To explore the limitation of CapsDecE2S, we collected and concluded the cases that CapsDecE2S fails to learn.

First, in all the failed cases, the top-10 failed words are the linking verbs, which include "{\textit{see}}", "{\textit{have}}", "{\textit{make}}", "{\textit{be}}", "{\textit{give}}", "{\textit{find}}", "{\textit{get}}", "{\textit{come}}", "{\textit{take}}" and "{\textit{feel}}". Usually, the linking verb connects the subject with a word that gives information about the subject, such as a condition or relationship. In most cases, the linking verbs do not describe any action, instead they link the subject with the rest of the sentence. It is hard for the CapsDecE2S model to learn the linking verb's true sense, especially since one word may occur in similar contexts. In fact, not only the CapsDecE2S model, most sense learning models are weak at these words.
Second, by random sampling 10\% of the failed cases, we find that except for the linking verbs, the majority are the words with quite close senses. Several typical examples are shown in Table~\ref{close_senses}.
The CapsDecE2S model mistakes one sense as the other and the weeny differences between the example sentences are hard to discover.

\section{Conclusion and Future Works}

In this paper, we have proposed to decompose the unsupervised word embedding with the capsule network and use the context and word sense matching training to learn the sense representation.
The experimental results on WiC and WSD datasets prove that the proposed CapsDecE2S method
contributes to learning more accurate sense than other compared methods.
These experiments show the potential of the information
contained in the unsupervised word embedding and also prove the
·feasibility of applying the capsule network to decompose unsupervised word embedding into context-specific sense representation.
Moreover, the analysis experiments show the enhanced interpretability of the capsule decomposing procedure and the context-specific sense representation.

The future works include 
1) merging the CapsDecE2S representation with other sense features to improve the sense learning ability;
2) exploring the diversity of words for sense learning where the words in SemCor3.0 corpus are greatly limited by the annotation cost;
3) applying the decomposed context-specific sense representation to downstream tasks;
4) proposing solid evaluation metrics to interpret the morpheme-like vectors and context-specific sense representation.



\section*{Acknowledgments}
This work is supported by Natural Science Foundation of China (Grant No. 61872113, 61573118, U1813215, 61876052), Special Foundation for Technology Research Program of Guangdong Province (Grant No. 2015B010131010), Strategic Emerging Industry Development Special Funds of Shenzhen (Grant No. JCYJ20170307150528934, JCYJ20170811153836555, JCYJ20180306172232154), Innovation Fund of Harbin Institute of Technology (Grant No. HIT.NSRIF.2017052)

\end{document}